\documentclass[12pt, final]{l4dc2020} 

\usepackage{times}

\usepackage{amsmath,amssymb,amsfonts}
\usepackage{algorithmic}
\usepackage{graphicx}
\usepackage{textcomp}
\usepackage{xcolor}
\def\BibTeX{{\rm B\kern-.05em{\sc i\kern-.025em b}\kern-.08em
    T\kern-.1667em\lower.7ex\hbox{E}\kern-.125emX}}
    
\usepackage{mathtools} 
\usepackage{hyperref}

\usepackage[justification=centering]{caption}
\usepackage[justification=centering, labelformat=empty]{subcaption}
\definecolor{burgundy}{rgb}{0.5, 0.0, 0.13}
\definecolor{ao(english)}{rgb}{0.0, 0.5, 0.0}

\usepackage{array}
\usepackage{booktabs}
\usepackage{adjustbox}
\newcolumntype{L}{>{\centering\arraybackslash}m{1.5cm}}
\newcolumntype{Y}{>{\centering\arraybackslash}X}
\newcolumntype{Z}{>{\centering\arraybackslash}X}

\usepackage[super]{nth}

\makeatletter
\newcommand{\setword}[2]{%
  \phantomsection
  #1\def\@currentlabel{\unexpanded{#1}}\label{#2}%
}
\makeatother

\DeclarePairedDelimiter\abs{\lvert}{\rvert}
\makeatletter
\let\oldabs\abs
\def\abs{\@ifstar{\oldabs}{\oldabs*}}

\let\originalleft\left
\let\originalright\right
\renewcommand{\left}{\mathopen{}\mathclose\bgroup\originalleft}
\renewcommand{\right}{\aftergroup\egroup\originalright}
\makeatletter
\def\resetMathstrut@{%
  \setbox\z@\hbox{%
    \mathchardef\@tempa\mathcode`\[\relax
    \def\@tempb##1"##2##3{\the\textfont"##3\char"}%
    \expandafter\@tempb\meaning\@tempa \relax
  }%
  \ht\Mathstrutbox@\ht\z@ \dp\Mathstrutbox@\dp\z@}

\let\oldforall\forall
 \renewcommand{\forall}{\ \oldforall \ }
 \let\oldexists\exists
 \renewcommand{\exists}{\ \oldexists \ }

\DeclareMathOperator*{\argmax}{\arg\!\max}

\usepackage{amssymb} 
\makeatletter
\newcommand*{\transpose}{%
  {\mathpalette\@transpose{}}%
}
\newcommand*{\@transpose}[2]{%
    \small \raisebox{1.37\depth}{$\m@th#1\intercal$}%
}
\makeatother
\newcommand*{\tran}{^{\mkern-1.5mu\transpose}}

\newcommand{\ie}{i\/.\/e\/.,\/~}
\newcommand{\eg}{e\/.\/g\/.,\/~}

\newtheorem{notations}{Notations}

\makeatletter
\def\N{\mathbb N}
\makeatletter
\newcommand*{\rom}[1]{\expandafter\@slowromancap\romannumeral #1@} \makeatother

\newcommand{\normal}[2]{\mathcal{N}(#1, #2)}

\newcommand{\given}{|}
\newcommand{\estimf}{\hat{f}}
\newcommand{\estimx}{\hat{x}}
\newcommand{\truef}{f}

\newcommand{\statespace}{\mathcal{X}}
\newcommand{\controlspace}{\mathcal{U}}

\newcommand{\postmean}{\mu}
\newcommand{\postvar}{\sigma^2}
\newcommand{\noise}{\epsilon}

\newcommand{\noisevar}{\sigma_\epsilon^2}
\newcommand{\entropy}{H}
\newcommand{\infogain}{I}

\title[Actively Learning Gaussian Process Dynamics]{Actively Learning Gaussian Process Dynamics}

\author{
 \Name{Mona Buisson-Fenet}$^{12}$ \Email{buissonfenet@is.mpg.de}\\
 \Name{Friedrich Solowjow}$^1$ \Email{solowjow@is.mpg.de}\\
  \Name{Sebastian Trimpe}$^1$ \Email{trimpe@is.mpg.de}\\
  \addr $^1$ Intelligent Control Systems Group, Max Planck Institute for Intelligent Systems, Stuttgart, Germany \\ 
  \addr $^2$ Centre Automatique et Syst\`{e}mes, MINES ParisTech, PSL University, Paris, France
}

\begin{document}


\maketitle

\vspace{-3mm}

\begin{abstract}
Despite the availability of ever more data enabled through modern sensor and computer technology, 
it still remains an open problem to learn dynamical systems in a sample-efficient way.
We propose active learning strategies that leverage information-theoretical properties arising naturally during Gaussian process regression while respecting constraints on the sampling process imposed by the system dynamics. 
Sample points are selected in regions with high uncertainty, leading to exploratory behavior and data-efficient training of the model.
All results are validated in an extensive numerical benchmark.
\end{abstract}



\section{Introduction}  \label{sec:intro}

Learning dynamical systems has received considerable attention over the last decades and is widely recognized as an important and difficult problem \citep{schon2011system}. 
Indeed, in the case of physical systems, sampling data often requires practically involved and time-consuming experiments.
Further, sampling at informative locations of the state space is challenging, since the system is constrained by the underlying dynamics. This is one key difference to many machine learning tasks, where data can be collected anywhere.
Hence, it is essential to excite the system in such a way that the generated data enables sample-efficient learning.
In the case of linear time-invariant (LTI) systems, there exists a rich body of theoretical results for this problem \citep{ljung2001system}. Nonetheless, it is still an active field of research, \eg see \citep{simchowitz2018learning} and references therein. 
Convergence results are usually tightly connected to the well-established theory of persistence of excitation \citep{Green_persistent_excitation_lin_sys}, which ensures that control inputs are significant enough to sufficiently excite the system. However, these control inputs are not necessarily optimal and targeted exploration can accelerate learning \citep{umenberger2019robust}.
Learning nonlinear systems is even more complex, although there have been many advances and progress over the years (see \eg \citep{schon2011system, Schoukens_survey_nonlin_sysid}).
We consider Gaussian process (GP) regression, which has been proven to be an efficient framework in many related applications,
including model learning \citep{Nguyen-Tuong_survey_model_learning_robot_control} or reinforcement learning \citep{Deisenroth_PILCO_presentation, Doerr_optimize_long-term_predictions_model-based_policy_search}.
These probabilistic models have many advantageous properties for learning dynamical systems \citep{Morari_learning_control_GP, Deisenroth_identify_GP_state_space_models, Doerr_probabilistic_recurrent_state-space_models}, such as taking uncertainty into account, coping with small datasets and incorporating prior knowledge.

Active learning, \ie sequentially choosing where to sample in order to build an informative dataset, has been investigated in many domains (see \citep{Charu_active_learning_survey} for an overview). 
A critical difference that sets the active learning problem for dynamical systems apart is the fact that it is not possible to arbitrarily sample the state-action space. 
Indeed, the system is constrained by the dynamics, and has to be excited appropriately by control inputs. 
Existing approaches for actively learning static maps thrive by incorporating information-theoretical criteria that guide the sampling procedure. 
In particular, the combination with GPs yields powerful theoretical and practical results \citep{Krause_sensor_placement_in_GP, Krause_nonmyopic_active_learning_GP}.
For dynamical systems, however, there is only little related work. 
Recent attempts have been made, proposing a greedy exploration scheme \citep{Morari_learning_control_GP}, focusing on exploration under safety constraints \citep{Koller_learning_MPC_safe_exploration, Fisac_safety_framework_learning_control_uncertain_robotics, Heim_learnable_safety_measure}, or on active exploration for reinforcement learning using linear Bayesian inference rather than GPs \citep{Belousov_receding_horizon_curiosity}. 
Approaches relying on parametrization of the trajectory have also been presented, including for learning time series with GPs \citep{Krause_informative_path_planning_underwater_vehicle, Nguyen-Tuong_safe_AL_time-series_GP}.
The proposed algorithms are related to this work, however, the analysis differs in several important points, which we will further discuss in Section \ref{sec:trajopt}.

\paragraph{Contributions}

We investigate the active learning problem for dynamical systems, which are modeled by a GP. 
In particular, we take the learned dynamics explicitly into account to guide the exploration.
The following contributions are made:
 \setlength\itemsep{0.01em}
\begin{itemize}
    \item Proposal of a method that searches for informative points to visit, then separately drives the system to reach them (\emph{separated search and control}, short \texttt{sep}). While we can provide some theoretical guarantees on the suboptimality of the sequence of locations to visit, we find the method to have limitations in practice.
	\item Novel method for deriving input trajectories by maximizing an information criterion, while taking the dynamics into account as constraints. We propose two variants, \emph{receding horizon} and \emph{plan and apply} (short \texttt{rec} and \texttt{p\&a}).
    \item Benchmark on a set of numerical experiments, including robotic systems from reinforcement learning benchmarks, showing the superiority of approaches based on joint optimization for actively exploring the state-action space. 
\end{itemize}


\section{Problem statement} \label{sec:problem_formulation}

We consider a system subject to the following discrete-time dynamics:
\begin{align}
x_{k+1} & = \truef(x_k, u_k)   \qquad y_k = x_k + \noise_k ,  \label{def_basic_dyn_syst}
\end{align}
where $\truef$ is an unknown Lipschitz-continuous function, $x_k \in \statespace$ is the system state at time step $k \in \mathbb{N}$, with $\statespace \subseteq \mathbb{R}^{d_x}$ the space of possible states, $u_k \in \controlspace$ is the control input, with $\controlspace \subseteq \mathbb{R}^{d_u}$ the space of bounded control inputs, and $\noise_k \sim \normal{0}{\noisevar}$ is i.i.d. Gaussian measurement noise. We assume the system has 
sufficient controllability and stability properties in order to exclude notoriously difficult learning problems, 
where systematic exploration would not be meaningful. 
For inputs $Z = (z_0, \cdots, z_n )\tran$, where $z_k = (x_k, u_k)$, we observe the noisy measurements $Y = (y_1, \cdots, y_{n+1}) \tran $. 
We are thus in the standard GP regression setting with noise-free input and noisy target.
While there are extensions to more realistic settings such as learning dynamics with noisy inputs \citep{Rasmussen_GP_training_input_noise_NIGP} and latent states \citep{Doerr_optimize_long-term_predictions_model-based_policy_search, Doerr_probabilistic_recurrent_state-space_models}, we do not consider them here for simplicity,  as they are orthogonal to the problem of excitation.
The true system dynamics $\truef$ are approximated by a GP denoted $\estimf$ (see \citep{Rasmussen_Williams_GP_for_ML} for an overview). It is fully characterized by its mean function $\postmean(\cdot)$ and its covariance function $k(\cdot,\cdot)$. The prediction $\estimf(z_*)$ at an unobserved point $z_*$ is normally distributed with posterior mean and variance
\begin{align}
\postmean(z_*) & = k_* \tran (K + \noisevar I)^{-1} Y 
\qquad
\postvar(z_*) = k_{**} - k_* \tran (K + \noisevar I)^{-1} k_* , \label{GP_post_expression}
\end{align}
where $K = (k(z_i, z_j))_{z_i, z_j \in Z}$ is the covariance matrix of $Z$, $k_* = (k(z_i, z_*))_{z_i \in Z} $, and $k_{**} = k(z_*, z_*)$. The prior mean is assumed to be zero without loss of generality. The differential entropy of $\estimf$ at $z_*$, which quantifies the uncertainty of the prediction \citep{MacKay_information_theory_inference_learning}, is defined as
$
\entropy(z_*) = \frac{1}{2} \log(2 \pi e \postvar(z_*)). \label{def_diff_entropy}
$
The kernel $k$ usually depends on some hyperparameters, which are optimized during learning, often by maximizing the data marginal log likelihood.

We address the following question: how should one excite \eqref{def_basic_dyn_syst} to generate samples $(Z, Y)$ for learning $\truef$ in a sample-efficient way? 
We derive control inputs that optimize information criteria such as differential entropy, while taking the autoregressive structure of \eqref{def_basic_dyn_syst} explicitly into account. Ultimately, this reduces the prediction error given a fixed number of samples, by choosing informative control inputs that lead to exploratory system behavior.

\section{From static to dynamic -- a fundamentally different problem} \label{sec:static_to_dyn}

Powerful active learning strategies have been developed for learning static maps \citep{Krause_sensor_placement_in_GP}. Here, it is possible to immediately query any point in the input space.
Thus, the problem is amenable to a clean information-theoretical treatment, which is lost for dynamical systems.
The insights from the well-studied static problem are a natural starting point for this work, and we present an extension to the dynamic setting herein.  At the same time, we shall underline the fundamentally different nature of the dynamical problem.

\subsection{The static problem: sensor placement} \label{subsec:sensor_placement}

A canonical example for actively learning static maps is the sensor placement problem \citep{Krause_sensor_placement_in_GP}.
The objective is to find the best locations of $N$ sensors $X = (x_1,...,x_N)$ out of a finite subset of possible locations, in order to approximate a static map $\truef$ with a GP $\estimf$, using noisy measurements $y_i = \truef(x_i) + \noise_i$. A possible solution is to select
$
X^{\mathrm{OPT}} = \argmax_{X} \infogain(X, \truef), 
$
where $\infogain(X,\truef) = \entropy(Y_X)-\entropy(Y_X \given \truef)$ is the mutual information between the observations $Y$ at $X$ and the underlying function $\truef$, and $\entropy$ is the differential entropy of the GP  $\estimf$. However, finding such an optimal set of placements is NP-hard \citep{Krause_sensor_placement_in_GP}. 
Therefore, there is a need for tractable approximations. In particular, the optimal sequence of placements can be approximated by the greedy rule
\begin{align}
x_i = \argmax_{x} \infogain(x \cup X_{i-1}, \truef) = \argmax_{x} \entropy_{i-1} (x) \quad \forall i \in \left\{1, ..., N\right\} , \label{sensor_placement_greedy_rule}
\end{align}
with $\entropy_{i}$ the differential entropy of the GP at iteration $i$.
Thanks to the submodularity and monotonicity of the function $\infogain(X, \truef)$ (see \citep{Krause_nonmyopic_nearoptimal_information_graphs, Krause_regret_bounds_GP_optimization_bandit} for details) and Proposition 4.3 in \citep{Nemhauser_approximations_submodular_functions}, it can be shown that the sequence $X^{\mathrm{G}}$ of greedy placements selected by \eqref{sensor_placement_greedy_rule} is close to the true optimal sequence $X^{\mathrm{OPT}}$:
\begin{align}
    \infogain(X^{\mathrm{G}}, \truef) \geq (1 - 1/e) \infogain(X^{\mathrm{OPT}}, \truef) .  \label{Nemhauser_theorem}
\end{align}

In this paper, we focus on differential entropy since we are considering a continuous space, for which the mutual information criterion proposed in \citep{Krause_sensor_placement_in_GP} cannot easily be computed.

\subsection{Extension to dynamical systems} \label{subsec:sep_approach_limits}

The dynamical problem is fundamentally different: we cannot sample at an arbitrary state $x$, we need to steer the system to $x$ through the unknown dynamics $f$ with a sequence of bounded actions $u$. 
Therefore, there is also an information gain along the trajectory, which is not considered in the previously introduced static framework.

At first, we ignore this fact: we separate the search for informative states from obtaining the control inputs that drive the system to these states. 
This method is denoted \emph{separated search and control} (\texttt{sep}).
Starting from an initial point $z_0 := (x_0, u_0)$, at each iteration the next location to visit is determined by the greedy rule
\begin{align}
z^\mathrm{G}_{i} = \argmax_{z \in \statespace \times \controlspace} \infogain(z \cup Z_{i-1}, \truef) = \argmax_{z \in \statespace \times \controlspace} \entropy_{i-1} (z) \quad \forall i \in \left\{1, ..., N\right\}, N \in \N. \label{submodular_greedy_procedure}
\end{align}
After solving \eqref{submodular_greedy_procedure}, we get a state-action pair $(x^\mathrm{G}_{i}, u^\mathrm{G}_{i})$. We steer the system to $x_i^G$ using a control trajectory $(u_k, ..., u_{k+M-1}) \in \controlspace ^{M}$, then apply $u^\mathrm{G}_{i}$, and update the GP and its hyperparameters with the data collected along the way.
Here, $M$ is the control horizon, and $k$ is the time step since the beginning of the experiment, while $i$ indexes the iterations of the greedy procedure. 
Due to the controllability assumption, the existence of such control trajectories is ensured for sufficiently large $M$ and there exist methods to obtain them, \eg iLQR  \citep{Mansard_iLQR_dynamic_programming, Tassa_iLQR_trajectory_optimization}. However, there are severe issues in the concrete implementation:

\begin{itemize}
 \setlength\itemsep{0.01em}
    \item In general, it is difficult to choose $M$ a priori such that each $z^\mathrm{G}_{i}$ is attainable in $M$ steps. However, limiting the search space in \eqref{submodular_greedy_procedure} to the points attainable in $M$ steps yields a time-varying set from which to choose $z$. In this case, Proposition 4.3 in  \citep{Nemhauser_approximations_submodular_functions} is not applicable anymore.
    Hence, suboptimality guarantees of type \eqref{Nemhauser_theorem} typically can only be derived for \eqref{submodular_greedy_procedure} if $z$ is chosen directly from $\statespace \times \controlspace$.
    \item With only the estimated dynamics $\estimf$ available for controller design, the location chosen by \eqref{submodular_greedy_procedure} might not actually be reached and the intended data point not be obtained.
    \item The considered search space $\statespace \times \controlspace$ is continuous, and not a finite set of possible locations as for the sensor placement problem. This has implications on the property of submodularity, which is originally defined for set functions.
    \item Solving \eqref{submodular_greedy_procedure}, a non-convex problem in a continuous state-action space, is nontrivial, and might not return the true optimum at each iteration.
\end{itemize}

One can obtain suboptimality guarantees of type \eqref{Nemhauser_theorem} for \eqref{submodular_greedy_procedure} under restrictive assumptions, namely: $\statespace \times \controlspace$ is a finite set, $z^\mathrm{G}_{i}$ is the true optimum at iteration $i$, and is actually visited by the control procedure.
However, these do not hold in practice. 
Information gain along the trajectory is also not included in this theoretical framework.
Next, we show that more efficient strategies can be designed by optimizing over the whole control trajectory.

\section{Informative control generation} \label{sec:trajopt}

Separating search from control yields an insightful embedding into the sensor placement problem. However, the crucial properties of this problem also become apparent, revealing that many aspects of the current solutions are not transferrable to dynamical systems. 
The discussed insufficiencies inspired us to jointly optimize for control inputs and informative states with respect to the approximate dynamics as constraints.  
We propose the following approach: at time step $k$, we pick the most informative control trajectory by solving 
\begin{align}
& U_k^* = \argmax_{(u_k,...,u_{k+M-1}) \in \controlspace ^M} \sum_{i=0}^{M-1} \entropy_k (\estimx_{k+i}, u_{k+i}) \label{trajopt_optim}
\\
& \text{s.t. } \estimx_{k+i+1} = \estimf_k(\estimx_{k+i}, u_{k+i}) , \ u_{k+i} \in \controlspace , \forall i \in \{0,...,M-1\} \notag
\end{align}
for a fixed time horizon $M$, $\estimx_k = x_k$, and $\estimx_{k+i+1}$ the mean of the GP prediction. 
This method is highly versatile: the cost function can easily be extended and further regularized, \eg by penalizing a suitable norm of the control signals.
Numerically, we use direct multiple shooting in CasADi \citep{Casadi} with Ipopt \citep{Ipopt}, as in \citep{Koller_learning_MPC_safe_exploration, Morari_learning_control_GP}. Note that \eqref{trajopt_optim} is only an upper bound of the entropy accumulated over the trajectory. This bound could be made sharper, for example by approximating the propagation of uncertainty through moment matching, but this necessitates added approximations and computations.

\paragraph{Receding horizon or plan and apply}

For the above-described method, we propose two options.
In the default setting, we update the GP and optimize its hyperparameters at each time step. Then, we solve \eqref{trajopt_optim} in a receding time fashion (variant denoted \emph{receding horizon}, short \texttt{rec}).
However, this is very costly in terms of computations and may lead to a shortsighted behavior, since the exploration strategy has a chance to ``change its mind'' every time step, which can lead to a greedy behavior if the optimization landscape is too flat.
Thus, we propose a computationally cheaper alternative: 
we solve \eqref{trajopt_optim}, roll out the whole control trajectory, batch update the GP with the measurements taken along the way, optimize its hyperparameters, and iterate (variant denoted \emph{plan and apply}, short \texttt{p\&a}). 
In this case, $M$ needs to be well-chosen: if it is too large, the GP will not be updated often enough, but if it is too small, the procedure will be too shortsighted.

\paragraph{Related algorithms}

Recent works propose related algorithms, but with a different focus. 
For example, in \citep{Nguyen-Tuong_safe_AL_time-series_GP}, exploration under safety constraints of dynamical systems modeled by GPs is investigated. However, exploration is achieved by parametrizing and selecting an informative piecewise linear trajectory in state space, which simplifies the optimization problem. 
An exploration scheme for learning GP dynamics is presented in \citep{Morari_learning_control_GP}, but it is greedy. 
The model learned during exploration is then used for reference tracking under uncertainty by solving an MPC problem. 
In \citep{Koller_learning_MPC_safe_exploration}, an MPC scheme for safe exploration of dynamical systems is proposed, where the constraint does not directly lie on the estimated dynamics, but on the propagation of safe ellipsoids through these dynamics. 
Exploration for reinforcement learning is considered in \citep{Belousov_receding_horizon_curiosity}, where a sequence of discrete actions is optimized. The proposed algorithm is related to \eqref{trajopt_optim}, but using linear Bayesian inference as the learning framework. In this paper, we focus on exploration for learning dynamical systems with GPs, with a novel viewpoint extending solutions of the static problem.  We also provide a comprehensive benchmark of control systems, which follows next.


\section{Numerical benchmark} \label{sec:benchmark}

\begin{figure}[t]
\centering
\subfigure[Pendulum]{\includegraphics[width=0.48\textwidth, height=0.2\textheight]{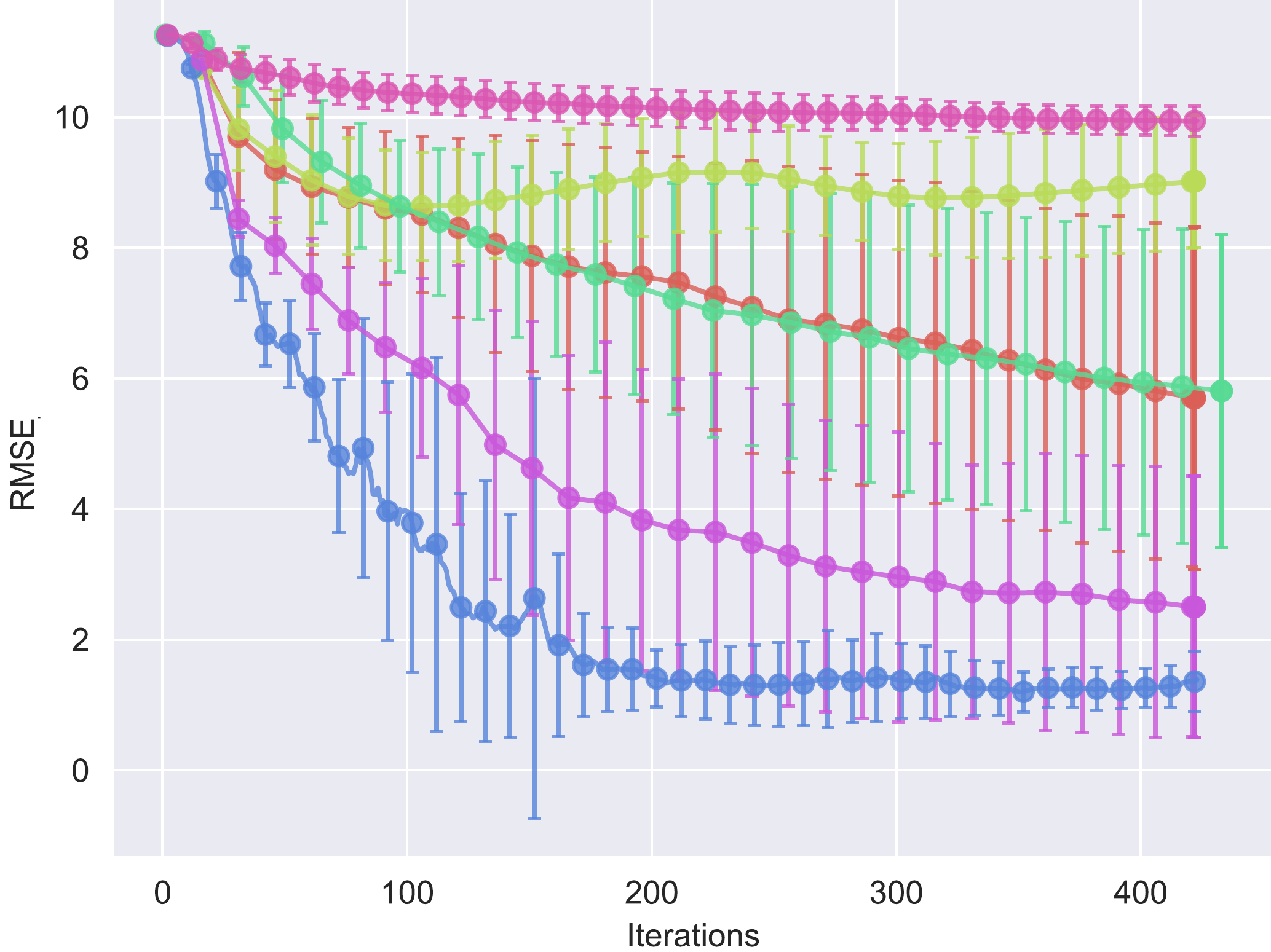}}\label{box_plot_pendulum}
\qquad
\hspace{-5mm}
\subfigure[Two-link robot]{\includegraphics[width=0.48\textwidth, height=0.2\textheight]{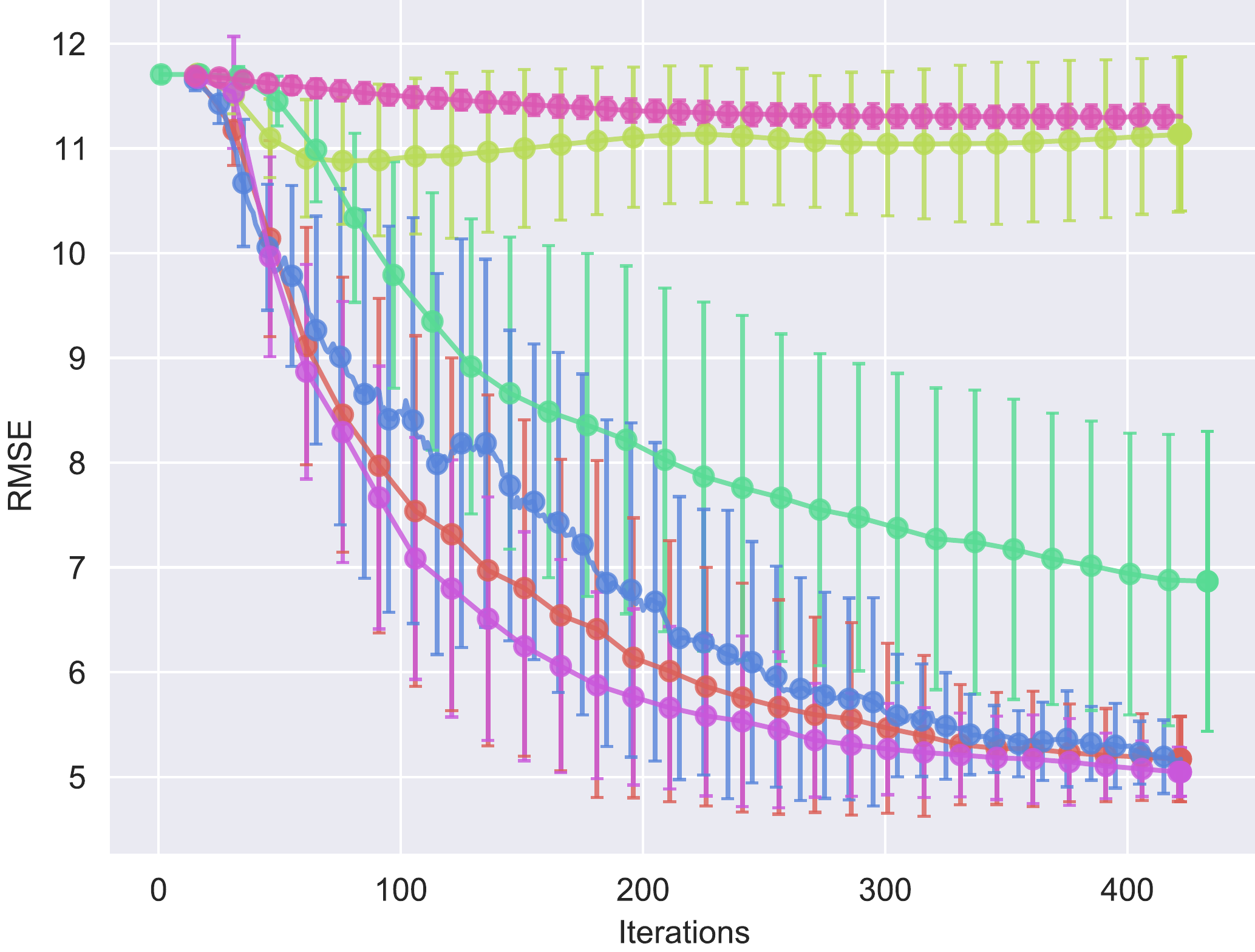}}\label{box_plot_2link}

\subfigure[DIPC]{\includegraphics[width=0.48\textwidth, height=0.2\textheight]{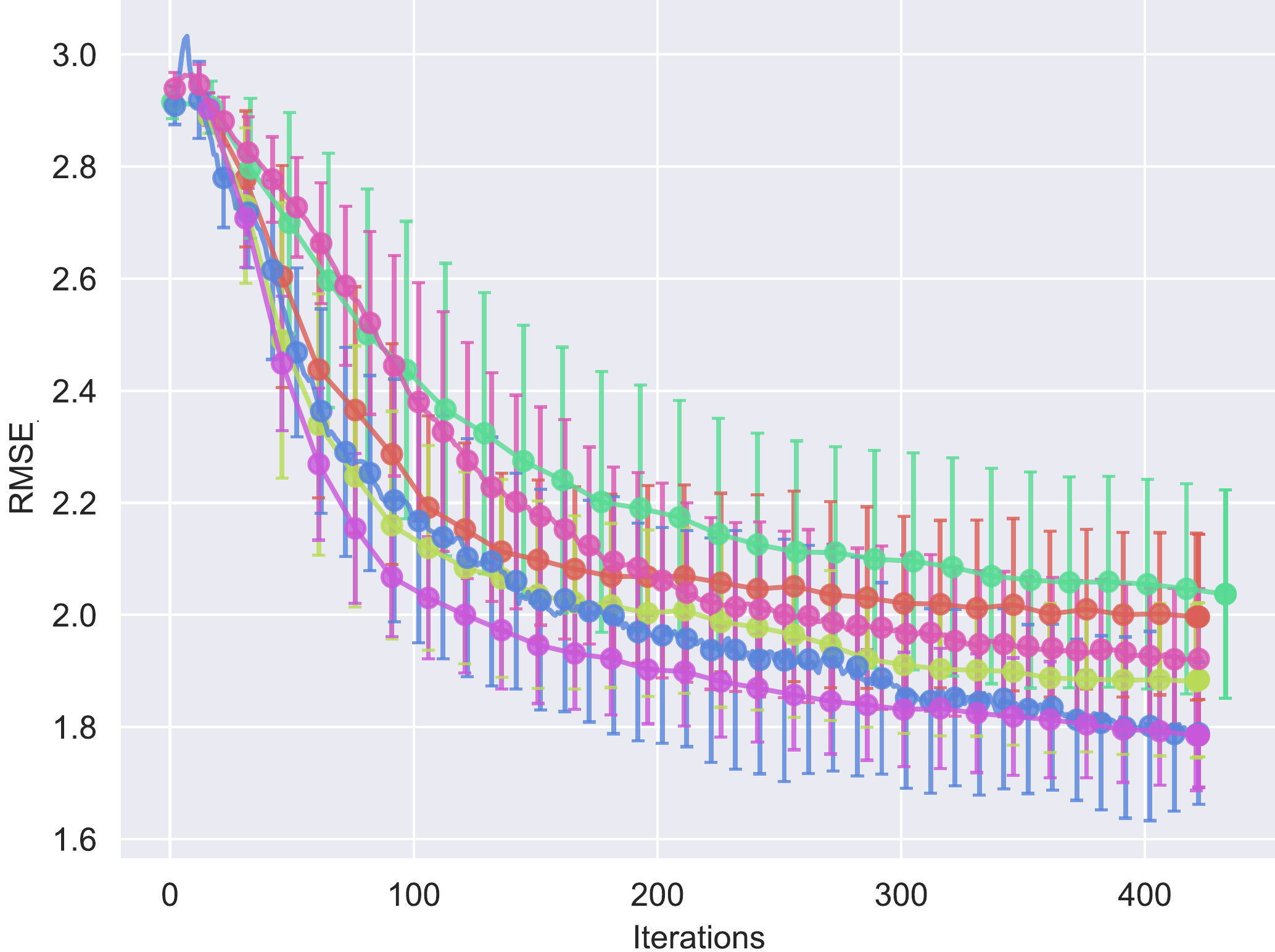}}\label{box_plot_DIPC}
\qquad
\hspace{-5mm}
\subfigure[Unicycle]{\includegraphics[width=0.48\textwidth, height=0.2\textheight]{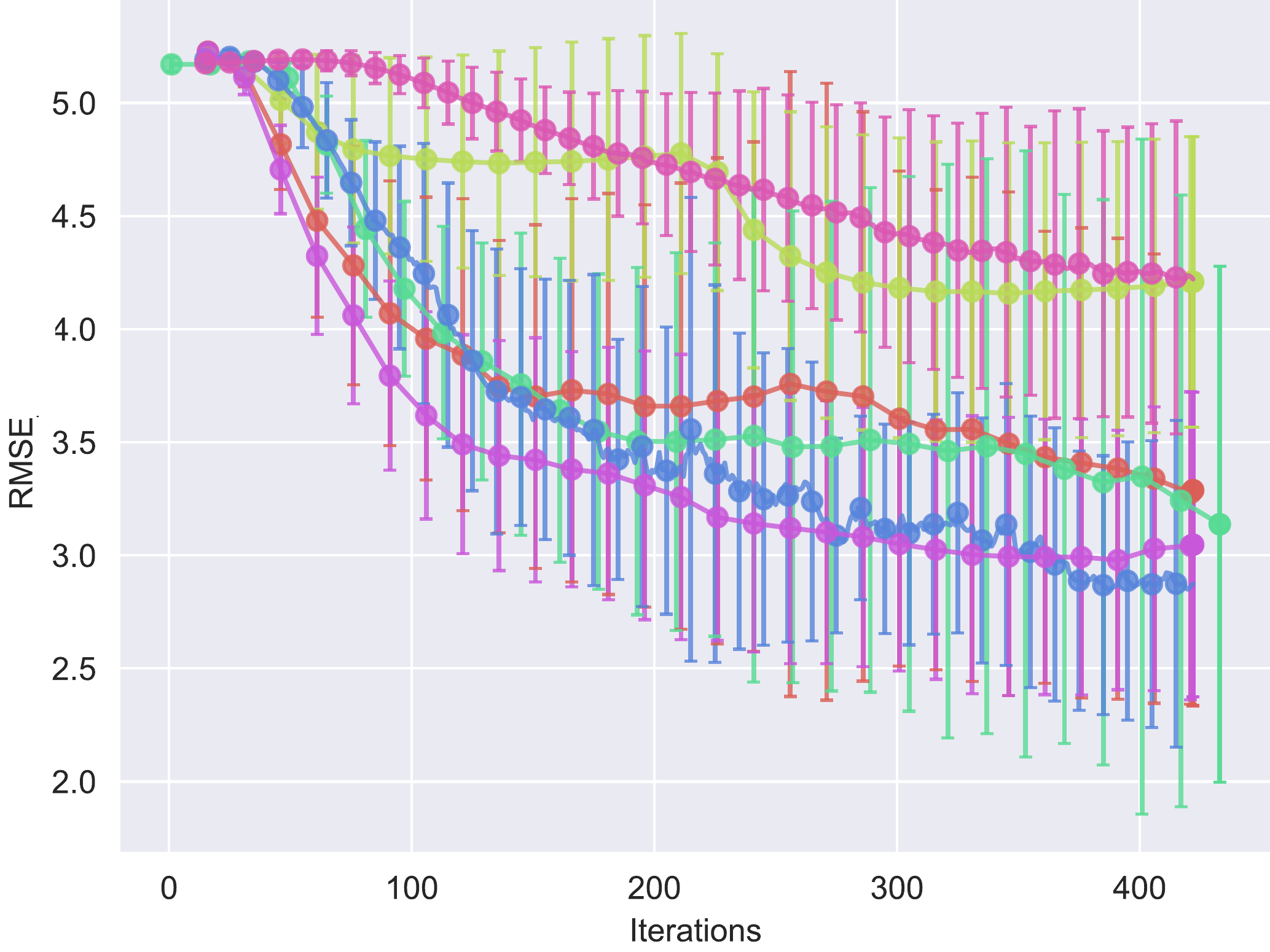}}\label{box_plot_unicycle}

\begin{minipage}[t]{0.5\linewidth}
	\centering
	\subfigure[Half-cheetah]{\includegraphics[width=\textwidth, height=0.2\textheight]{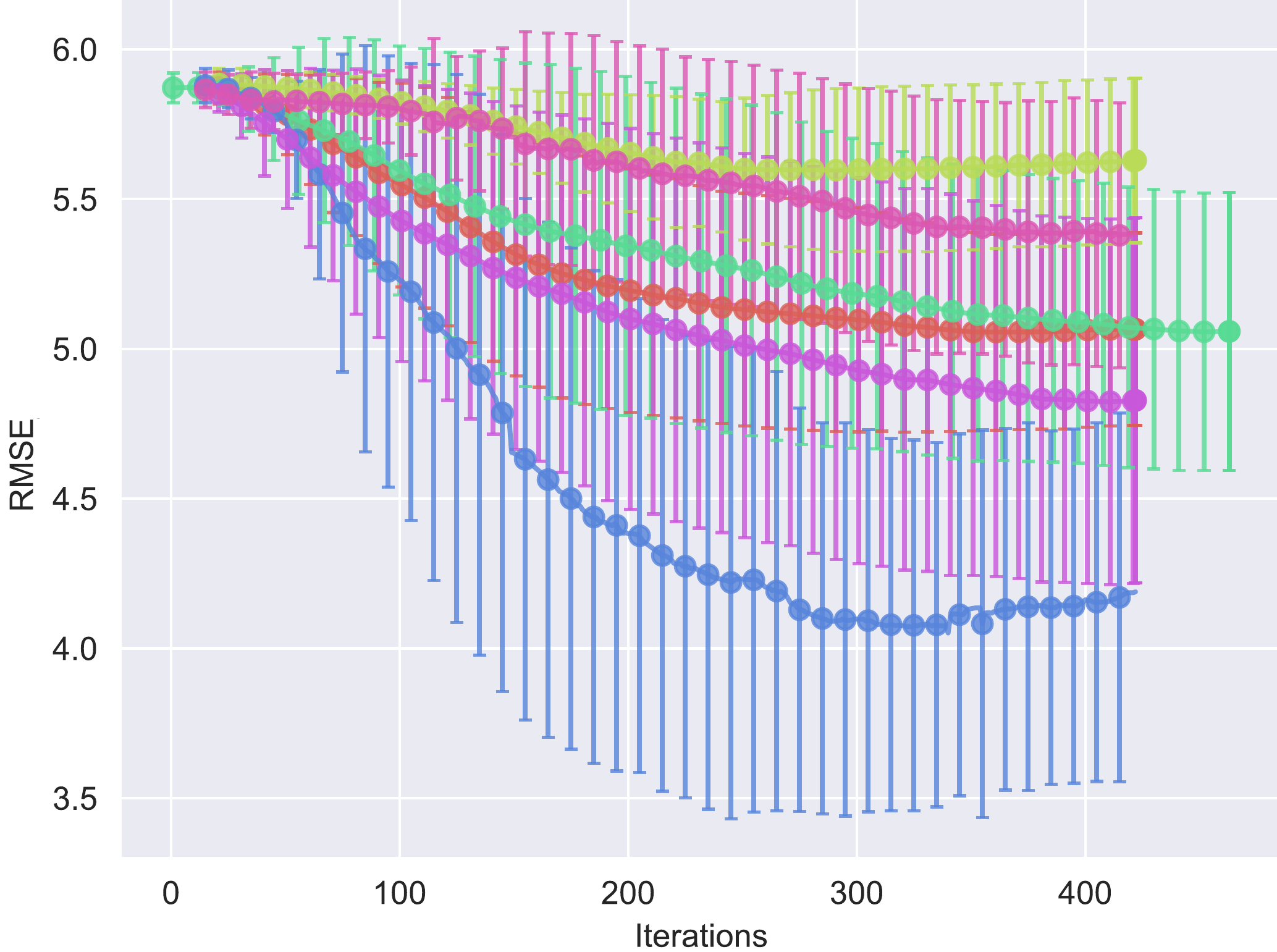}}\label{box_plot_cheetah}
\end{minipage}
\qquad
\qquad
\begin{minipage}[t]{0.1\linewidth}
	\centering
	\vspace{2.6mm}
	\includegraphics[width=\textwidth, height=0.18\textheight]{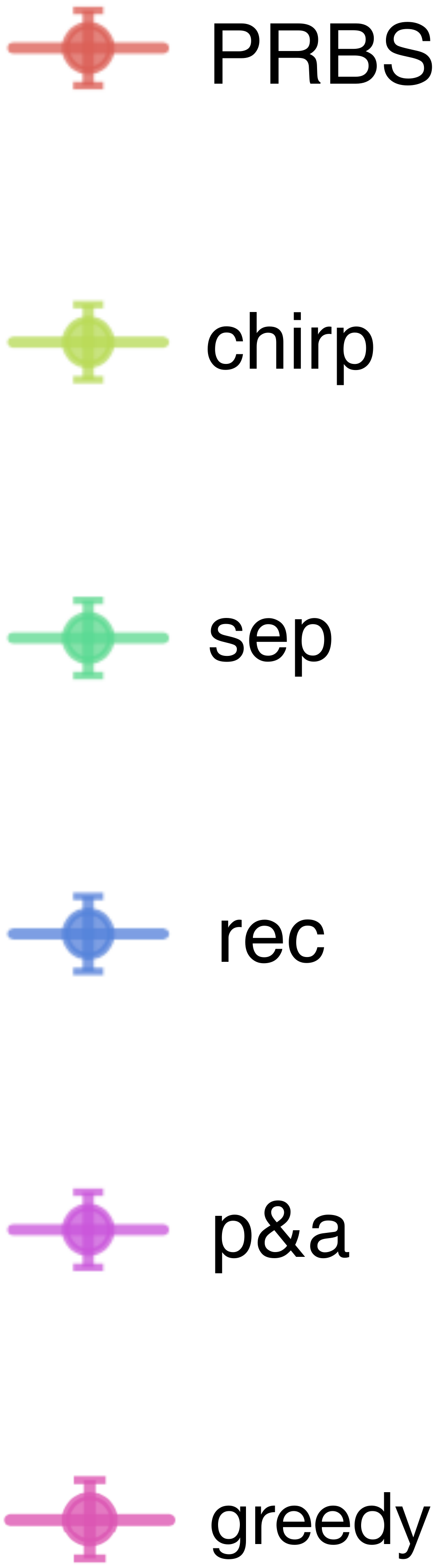}\label{box_plot_legend}
\end{minipage}
\qquad
\hspace{-13mm}
\begin{minipage}[t][.20\textheight]{0.25\linewidth}
	\centering
	\vspace{1.1mm}
	\includegraphics[width=0.7\textwidth, height=0.03\textheight]{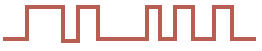}
	\vfill
	\includegraphics[width=0.7\textwidth, height=0.03\textheight]{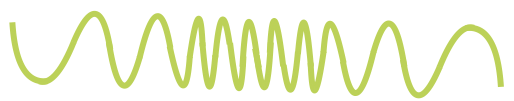}
	\vfill
	\includegraphics[width=0.5\textwidth, height=0.03\textheight]{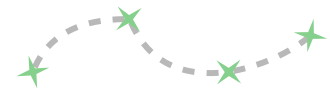}
	\vfill
	\includegraphics[width=0.5\textwidth, height=0.03\textheight]{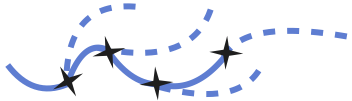}
	\vfill
	\includegraphics[width=0.5\textwidth, height=0.03\textheight]{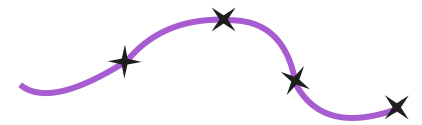}
	\vfill
	\includegraphics[width=0.5\textwidth, height=0.03\textheight]{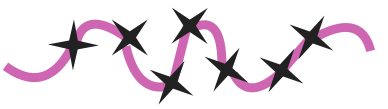}
\end{minipage}

\caption[Box plots for pendulum simulations]{RMSE over time for all benchmark systems (mean $\pm$ standard deviation).}%
\label{table_box_plots}
\end{figure}

We compare the proposed methods in a numerical benchmark. 
For each approach, we evaluate the prediction error and quantify how much of the state space has been explored.
The results are summarized in Table \ref{table_benchmark_results_summary} and Figure \ref{table_box_plots}.

\begin{figure}[t]
\centering
\subfigure[Pendulum]{\includegraphics[width=0.12\textwidth, height=0.1\textheight]{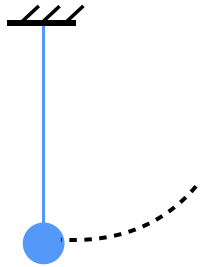}}\label{illustration_pendulum}
\qquad
\hspace{-2mm}
\subfigure[Two-link robot]{\includegraphics[width=0.18\textwidth, height=0.1\textheight]{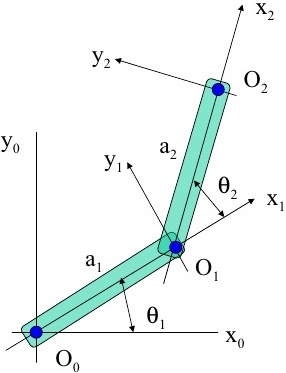}}\label{illustration_2link}
\qquad
\hspace{-7mm}
\subfigure[DIPC]{\includegraphics[width=0.19\textwidth, height=0.1\textheight]{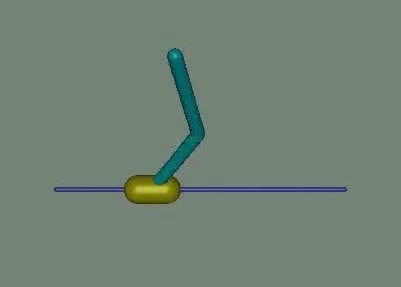}}\label{illustration_DIPC}
\qquad
\hspace{-7mm}
\subfigure[Unicycle]{\includegraphics[width=0.19\textwidth, height=0.1\textheight]{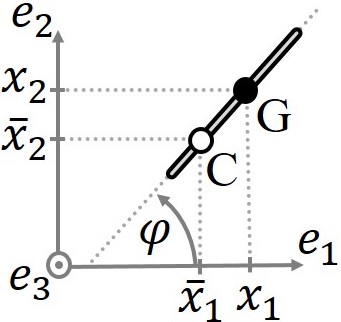}}\label{illustration_unicycle}
\qquad
\hspace{-7mm}
\subfigure[Half-cheetah]{\includegraphics[width=0.19\textwidth, height=0.1\textheight]{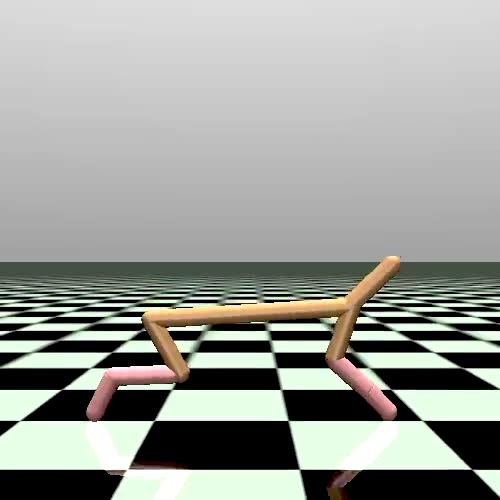}}\label{illustration_halfcheetah}

\caption[Illustration systems]{Illustration of the benchmark systems.}%
\label{systems_illustrated}
\end{figure}

\paragraph{Experimental settings}
We run the following methods\footnote{Code available at \url{https://git-amd.tuebingen.mpg.de/mbuissonfenet/active_learning_gp.git}}: pseudorandom binary sequences (PRBS) and chirps (\citep{Nelles_nonlin_sysid}, Section 17.7) to compare to standard system identification signals, separated search and control (\texttt{sep}), and our method based on optimal control \eqref{trajopt_optim} with either receding horizon (\texttt{rec}), plan and apply (\texttt{p\&a}). We also compare with the greedy strategy $(M=1)$. 
Random control inputs with a holding time of either $M$ or 1 were also investigated, but they did not challenge the other standard signals and are not included in the final plots.
We focus on the standard GP setting: squared exponential kernel, independent GPs for each output dimension, hyperparameters that maximize the marginal log likelihood.
We evaluate on five nonlinear dynamical systems with continuous state-action space and bounded controls, illustrated in Figure \ref{systems_illustrated}: 
\begin{itemize}
\setlength\itemsep{0.2em}
    \item Pendulum, with $d_x=2$, $d_u=1$, $\noisevar = 0.05$, $M=15$;
    \item Two-link planar robot \citep{Siciliano_robotics_book}, with $d_x=4$, $d_u=2$, $\noisevar = 0.05$, $M=15$;
    \item Double inverted pendulum on a cart (DIPC) from the MuJoCo environment \citep{Todorov_MuJoCo_presentation} in Gym \citep{Brockman_OpenAI_Gym_presentation}, with $d_x=8$, $d_u=1$, $\noisevar = 0.05$, $M=15$, and added damping;
    \item Unicycle \citep{udwadia2007analytical}, with $d_x=6$, $d_u=2$, $\noisevar=0.001$, $M=15$;
    \item Half-cheetah, also from MuJoCo in Gym, with $d_x=18$, $d_u=6$, $\noisevar=0.001$, $M=10$.
\end{itemize}

Each method uses the same number of data points and the same planning horizon (except for the greedy method which has a horizon of 1) and starts from the same stable equilibrium. We make sure each system has enough damping to be sufficiently stable and controllable in the exploration region, and choose the bounds on $\controlspace$ such that exploring the state space is neither too easy (even random signals can easily go everywhere) nor too hard (even active exploration methods cannot go far). We run 10 trials of \texttt{rec} since it is computationally heavy, and 100 trials of all other methods.

\paragraph{Evaluation criteria}

We quantify the accuracy of the learned model by monitoring the root mean square prediction error (RMSE) over a grid of uniformly randomly distributed states and inputs in a predefined region of interest, and the quality of exploration by computing the percentage of coverage of this region at the end of the experiment. The region of interest is chosen a priori in $\statespace$ for each system, by heuristically picking bounds in each dimension in which most of our experiments stay when started from the same stable equilibrium. The percentage of coverage is computed by discretizing this region and computing the average number of cells visited during the experiment. In this paper, we are not as much interested in the absolute results but rather in the comparison between the different methods, and in demonstrating that some explore the state space more than others, yielding a more accurate model.

\paragraph{Results}

The results shown in Table \ref{table_benchmark_results_summary} and Figure \ref{table_box_plots} confirm that our optimization-based exploration methods (\texttt{rec} and \texttt{p\&a}) yield the lowest prediction error and thus, the best models. 
Indeed, they can push each nonlinear system to unknown regions of $\statespace \times \controlspace$, generating informative data points in the whole state space, which yields an overall more consistent model.
The \texttt{sep} method performs reasonably well, but significantly worse than \texttt{rec} and \texttt{p\&a}. 
The greedy method is not able to explore as much of the state space as it is too shortsighted.
The standard signals from system identification do not consider the current model and therefore, can perform arbitrarily badly.
Nonetheless, PRBS explored surprisingly well for systems that are close to linear, \eg rigid-body dynamics with torque control, where several states are linear in the input. However, this is not the case for other types of systems (\eg DIPC), and PRBS is often not a desirable system behavior.

\begin{table*}[t]
\caption[Comparison of benchmark results]{Final RMSE and coverage of the state space (mean $\pm$ standard deviation).}
\label{table_benchmark_results_summary}
\footnotesize
\tabcolsep=0.11cm
\begin{tabular*}{\textwidth}{@{}l @{\extracolsep{\fill}}@{\hskip -2in} *{5}{c} @{}}
\toprule
 & Pendulum & Two-link & DIPC & Unicycle & Half-cheetah \\
\toprule
Final RMSE &  &  &  &  &  \\
\bottomrule
PRBS & $5.7 \pm 2.6$ &  $5.2 \pm 0.4$  & $ 2.00 \pm 0.15 $  & $ 3.3 \pm 1.0 $  & $ 5.1 \pm 0.3 $  \\
chirp & $9.0 \pm 1.0$ &  $11.1 \pm 0.7$  & $1.89 \pm 0.14 $  & $ 4.2 \pm 0.6 $  &  $ 5.6 \pm 0.3 $ \\
sep & $5.8 \pm 2.4$ &  $6.9 \pm 1.4 $  & $ 2.03 \pm 0.19 $  & $ 3.1 \pm 1.1 $  & $ 5.1 \pm 0.5 $  \\
rec & {\color{ao(english)}$1.4 \pm 0.5$} &  $5.2 \pm 0.4 $  &  {\color{ao(english)}$ 1.79 \pm 0.12 $} &  {\color{ao(english)}$ 2.9 \pm 0.7 $}  &  {\color{ao(english)}$ 4.2 \pm 0.6 $}  \\
p\&a & $2.5 \pm 2.0$ &  {\color{ao(english)}$5.0 \pm 0.2 $}  & {\color{ao(english)}$1.79 \pm 0.09 $}  & $3.0 \pm 0.7 $  & $ 4.8 \pm 0.6 $  \\
greedy & $9.9 \pm 0.2$ &  $11.3 \pm 0.1$  & $ 1.92 \pm 0.13 $  & $ 4.2 \pm 0.7 $  & $ 5.4 \pm 0.4 $  \\
\bottomrule
Coverage of state space (in percent of the region of interest) &  &  &  &  &  \\
\midrule
PRBS & $22.1 \pm 11.0 $ &  $52.8 \pm 7.6  $ &  $ 67.4 \pm 8.2 $ & $ 65.8 \pm 7.3 $  & $50.7 \pm 4.9 $  \\
chirp & $14.0 \pm 2.6 $ &  $20.0 \pm 6.3 $ & $ 72.4 \pm 9.0 $  & $ 41.9 \pm 13.8 $  & $54.4 \pm 9.0 $  \\
sep & $22.0 \pm 9.1 $ &  $39.7 \pm 7.3 $  & $ 64.4 \pm 7.5  $  & $ 66.0 \pm 7.1  $ & $57.2 \pm 4.3 $  \\
rec & {\color{ao(english)}$46.4 \pm  6.6 $} &  {\color{ao(english)}$53.2 \pm 7.7 $}  & $ 74.8 \pm 6.5 $  & $ 65.6 \pm 5.2 $  &   {\color{ao(english)}$69.1 \pm 3.1 $} \\
p\&a & $37.1 \pm 9.0 $ &  $51.6 \pm 7.5 $  &  {\color{ao(english)}$ 75.9 \pm 5.3  $} &  {\color{ao(english)} $66.1 \pm 5.9  $}  & $65.3 \pm 3.8 $  \\
greedy & $10.5 \pm 1.0$ &  $16.4 \pm 1.1$  & $ 68.2 \pm 4.6 $  & $ 51.9 \pm 5.3 $  & $ 67.1 \pm 3.2 $  \\
\bottomrule
\end{tabular*}
\end{table*}

\section{Conclusion}  \label{sec:conclusion}

When learning models of dynamical systems, efficient exploration is key as it determines how informative the collected data is.
In this paper, we propose and benchmark three main algorithms for actively learning dynamical systems with GPs.
The separated search and control method is inspired by active learning for static GPs.
However, its performance is suboptimal, and the theoretical guarantees of the static case are not directly applicable. 
More efficient exploration can be obtained by computing optimal excitation signals with respect to an information criterion.
The receding horizon variant of this approach performs well but yields a high computational burden. Hence, we also propose a batch update that trades off computation against performance.
This framework is efficient but also versatile, as further modifications of the cost function are straightforward. 
We show on a numerical benchmark of diverse dynamical systems that the proposed methods are capable of exploring the state-action space efficiently, yielding more informative data, and hence, a more accurate model. 
In future work, we intend to study the effects of different cost functions and to generalize our framework to more realistic GP models, for example with noisy inputs and latent states. 
Validation on hardware experiments would also be relevant, however, the methods need to be computationally optimized first.
 


\acks{This work was supported in part by the Cyber Valley Initiative, the International Max Planck Research School for Intelligent Systems, and the Max Planck Society.}

\bibliography{L4DC2020_active_learning_dyn_GP.bib}

\end{document}